\documentclass{article}

\usepackage[numbers]{natbib}
\usepackage{arxiv}
\usepackage{graphics}
\usepackage{epsfig}
\usepackage{mathptmx}
\usepackage{times}
\usepackage{amsmath}
\usepackage{amssymb}
\usepackage{caption}
\usepackage{subfig}
\usepackage{xspace}
\usepackage{algorithm}
\usepackage{algpseudocode}
\usepackage{tikz}
\usepackage{tikzscale}
\usepackage{pgfplots}
\pgfplotsset{compat=1.12}
\usetikzlibrary{matrix}
\usepgfplotslibrary{groupplots}
\pgfplotsset{compat=newest}
\newcommand{\eg}{\emph{e.g.}\@\xspace}
\DeclareMathOperator{\atan}{atan}

\let\vecbold\mathbf
\usepackage{stmaryrd}
\usepackage{float}

\title{\LARGE \bf
Coupled Longitudinal and Lateral Control of a Vehicle\\ using Deep Learning\thanks{Published in the IEEE 2018 International Conference on Intelligent Transportation Systems (ITSC 2018). This work was supported by the international Chair MINES ParisTech - Peugeot-Citro\"en - Safran - Valeo on ground vehicle automation.   (*) Guillaume Devineau and Philip Polack contributed equally to this work. (**) Florent Altch\'e is also with \'Ecole des Ponts ParisTech, Cit\'e Descartes, 6-8 Av Blaise Pascal, 77455 Champs-sur-Marne, France. }%
}

\author{Guillaume Devineau$^{\star}$\\
  Center for Robotics, MINES ParisTech\\PSL Research University\\Paris, France\\
  \texttt{guillaume.devineau@mines-paristech.fr}  
\And Philip Polack$^{\star}$\\
  Center for Robotics, MINES ParisTech\\PSL Research University\\Paris, France\\
  \texttt{philip.polack@mines-paristech.fr}  
\And Florent Altch\'e$^{**}$\\
  Center for Robotics, MINES ParisTech\\PSL Research University\\Paris, France\\
  \texttt{florent.altche@mines-paristech.fr}  
\And Fabien Moutarde\\
  Center for Robotics, MINES ParisTech\\PSL Research University\\Paris, France\\
  \texttt{fabien.moutarde@mines-paristech.fr}
}

\begin{document}

\maketitle

\begin{abstract}
This paper explores the capability of deep neural networks to capture key characteristics of vehicle dynamics, and their ability to perform coupled longitudinal and lateral control of a vehicle. To this extent, two different artificial neural networks are trained to compute vehicle controls corresponding to a reference trajectory, using a dataset based on high-fidelity simulations of vehicle dynamics. In this study, control inputs are chosen as the steering angle of the front wheels, and the applied torque on each wheel. The performance of both models, namely a Multi-Layer Perceptron (MLP) and a Convolutional Neural Network (CNN), is evaluated based on their ability to drive the vehicle on a challenging test track, shifting between long straight lines and tight curves. A comparison to conventional decoupled controllers on the same track is also provided.
\end{abstract}

\section{INTRODUCTION}
\label{sec:introduction}
The recent development of deep learning has led to dramatic progress in multiple research fields, and this technique has naturally found applications in autonomous vehicles. The use of deep learning to perform perceptive tasks such as image segmentation has been widely researched in the last few years, and highly efficient neural network architectures are now available for such tasks. More recently, several teams have proposed taking deep learning a step further, by training so-called ``end-to-end'' algorithms to directly output vehicle controls from raw sensor data (see, in particular, the seminal work in \cite{Bojarski2016}).

Although end-to-end driving is highly appealing, as it removes the need to design motion planning and control algorithms by hand, handing the safety of the car occupants to a software operating as a black box seems problematic. A possible workaround to this downside is to use ``forensics'' techniques that can, to a certain extent, help understand the behavior of deep neural networks \cite{Castelvecchi2016}.

We choose a different approach consisting in breaking down complexity by training simpler, mono-task neural networks to solve specific problems arising in autonomous driving; we argue that the reduced complexity of individual tasks allows much easier testing and validation.

In this article, we focus on the problem of controlling a car-like vehicle in highly dynamic situations, for instance to perform evasive maneuvers in face of an obstacle. A particular challenge in such scenarios is the important coupling between longitudinal and lateral dynamics when nearing the vehicle's handling limits, which requires highly detailed models to properly take into account \cite{Gillespie1997}.
However, precisely modeling this coupling involves complex non-linear relations between state variables, and using the resulting model is usually too costly for real-time applications. For this reason, most references in the field of motion planning mainly focus on simpler models, such as point-mass or kinematic bicycle (single track), which are constrained to avoid highly coupled dynamics \cite{PolackACC2018}.
Similarly, research on automotive control usually treats the longitudinal and lateral dynamics separately in order to simplify the problem \cite{Khodayari2010}.

Although these simplifications can yield good results in standard on-road driving situations, they may be problematic for vehicle safety when driving near its handling limits, for instance at high speed or on slippery roads. To handle such situations, some authors have proposed using Model Predictive Control (MPC) with a simplified, coupled dynamic model~\cite{Falcone2007a} which is limited to extremely short time horizons (a few dozen milliseconds) to allow real-time computation. Other authors have proposed to model the coupling between longitudinal and lateral motions using the concept of ``friction circle'' \cite{Kritayakirana2012a}, 
which allows precisely stabilizing a vehicle in circular drifts~\cite{Goh2016}. However, the transition towards the stabilized drifting phase -- which is critical in the ability, \eg, to perform evasive maneuvers -- remains problematic with this framework.

In this article, we propose to use deep neural networks to implicitly model highly coupled vehicular dynamics, and perform low-level control in real-time. In order to do so, we train a deep neural network to output low-level controls (wheels torque and steering angle) corresponding to a given initial vehicle state and target trajectory. Compared to classical MPC frameworks which require integrating dynamic equations on-line, this approach allows to perform this task off-line and use only simple mathematical operations on-line, leading to much faster computations.

Several authors have already proposed a divide-and-conquer approach by using machine learning on specific sub-tasks instead of performing end-to-end computations, and in particular on the case of motion planning and control. For instance, reference~\cite{Drews2017} used a Convolutional Neural Network (CNN) to generate a cost function from input images, which is then used inside an MPC framework for high-speed autonomous driving; however, this approach has the same limitations as model predictive control. Other approaches, such as \cite{Se-YoungOh2000}, used reinforcement learning to output steering controls for a vehicle, but were limited to low-speed applications. Reference~\cite{Punjani2015} used a Rectified Linear Unit (ReLU) network model to identify the dynamics of a helicopter in order to predict its future accelerations, but this model has not been used for control. 

Closer to our work, reference \cite{Rivals1994} trained neural networks integrating a priori knowledge of the bicycle model for decoupled longitudinal and lateral control of a vehicle; in \cite{Chen2017}, authors used supervised learning to generate lateral controls for truck and integrated a control barrier function to ensure the safety of the system. Reference \cite{Cui2017} coupled a standard control and an adaptive neural network to compensate for unknown perturbations in order to perform trajectory tracking for autonomous underwater vehicle. To the best of our knowledge, deep neural networks have not been used in the literature for the coupled control of wheeled vehicles.

The rest of this article is organized as follows: Section~\ref{sec:vehicle_model} presents the vehicle model used to generate the training dataset and to simulate the vehicle dynamics on a test track. Section~\ref{sec:DL_models} introduces two artificial neural networks architectures used to generate the control signals for a given target trajectory, and describes the training procedure used in this article. Section~\ref{sec:results} compares the performance of these two networks, using simulation on a challenging test track. A comparison to conventional decoupled controllers is also provided. Finally, Section~\ref{sec:conclusions} concludes this study.

\section{THE 9~DoF VEHICLE MODEL}
\label{sec:vehicle_model}
In this section, we present the 9 Degrees of Freedom (9~DoF) vehicle model which is used both to generate the training and testing dataset, and as a simulation model to evaluate the performance of the deep-learning-based controllers.

\begin{figure}[h!]
	\centering
	\includegraphics[scale=0.4]{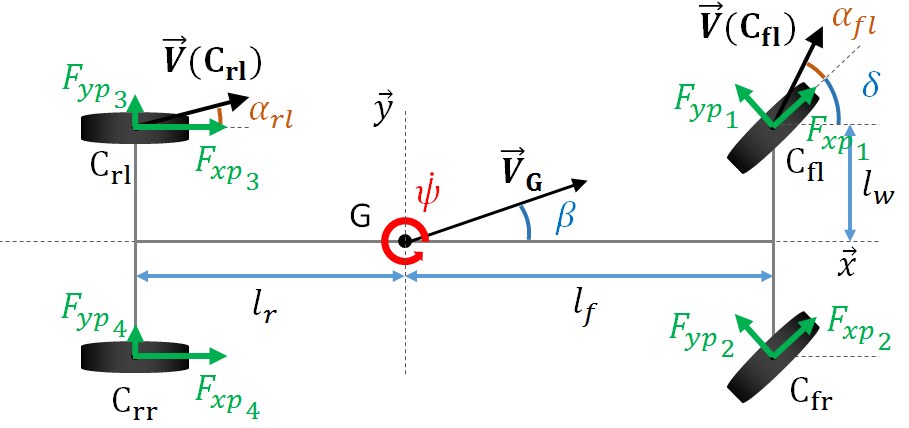}       
	\caption{Vehicle model and notations.}
	\label{fig:carSim}
\end{figure}

The Degrees of Freedom comprise 
3~DoF for the vehicle's motion in a plane ($V_x, V_y, \dot{\psi}$), 
2~DoF for the carbody's rotation ($\dot{\theta}, \dot{\phi}$)
and 4~DoF for the rotational speed of each wheel ($\omega_{fl},\omega_{fr},\omega_{rl},\omega_{rr}$). 
The model takes into account both the coupling of longitudinal and lateral slips and the load transfer between tires. The control inputs of the model are the torques $T_{\omega_i}$ applied at each wheel $i$ and the steering angle $\delta$ of the front wheel. The low-level dynamics of the engine and brakes are not considered here. The notations are given in Table \ref{tab:notations} and illustrated in Figure~\ref{fig:carSim}.

\textit{Remark: }the subscript $i=1..4$ refers respectively to the front left ($fl$), front right ($fr$), rear left ($rl$) and rear right ($rr$) wheels.

Several assumptions were made for the model:
\begin{itemize}
	\item Only the front wheels are steerable.
	\item The roll and pitch rotations happen around the center of gravity.
	\item The aerodynamic force is applied at the height of the center of gravity. Therefore, it does not involve any moment on the vehicle.
	\item The slope and road-bank angle of the road are not taken into account.
\end{itemize}

\begin{table}[h]
	\vspace{+0.08in}
	\caption{Notations}
	\label{tab:notations}
	\begin{tabular}{p{2.4cm} p{12.8cm}}
		\hline
		\\
		$X$, $Y$ & Position of the vehicle in the ground frame\\
		$\theta$, $\phi$, $\psi$ & Roll, pitch and yaw angles of the carbody \\
		$V_x$, $V_y$ & Longitudinal and lateral speed of the vehicle in its inertial frame \\
		$M_T$ & Total mass of the vehicle\\
		$I_x$, $I_y$, $I_z$ & Inertia of the vehicle around its roll, pitch and yaw axis\\
		$I_{r_i}$ & Inertia of the wheel $i$\\
		$T_{\omega_i}$ & Total torque applied to the wheel $i$\\
		$F_{xp_i}$, $F_{yp_i}$  & Longitudinal and lateral tire forces generated by the road on the wheel $i$ expressed in the tire frame\\
		$F_{x_i}$, $F_{y_i}$ & Longitudinal and lateral tire forces generated by the road on the wheel $i$ expressed in the vehicle frame  $(x,y)$\\
		$F_{z_i}$ & Normal reaction forces on wheel $i$\\
		$l_f$, $l_r$ & Distance between the front (resp. rear) axle and the center of gravity\\
		$l_w$ & Half-track of the vehicle\\ 
		$h$ & Height of the center of gravity\\
		$r_{eff}$ & Effective radius of the wheel\\ 
		$\omega_i$ & Angular velocity of the wheel $i$ \\
		$V_{xp_i}$ & Longitudinal speed of the center of rotation of wheel $i$ expressed in the tire frame\\
		\\
		\hline
	\end{tabular}
\end{table}

\subsection{Vehicle dynamics}
\label{ssec:CarModel}

Equations~(\ref{eq:veh_dyn_1}-\ref{eq:veh_dyn_5}) give the expression of the vehicle dynamics:
\begin{subequations}
	\label{eq:vehicle_dynamics}
	\begin{eqnarray}
	\label{eq:veh_dyn_1}
	M_T \dot{V}_x & = & M_T\dot{\psi} V_y + \sum_{i=1}^4 F_{x_i} - F_{aero}\\
	\label{eq:veh_dyn_2}
	M_T\dot{V}_y & = & - M_T\dot{\psi} V_x +  \sum_{i=1}^4 F_{y_i}\\
	\label{eq:veh_dyn_3}
	I_x\ddot{\theta} & = & l_w (F_{z_1}+F_{z_3}-F_{z_2}-F_{z_4}) + h \sum_{i=1}^4 F_{y_i}\\
	\label{eq:veh_dyn_4}
	I_y\ddot{\phi} & = & l_r (F_{z_3} + F_{z_4}) -l_f (F_{z_1} +F_{z_2}) - h \sum_{i=1}^4 F_{x_i}\\
	\label{eq:veh_dyn_5}
	I_z\ddot{\psi} & = & l_f (F_{y_1} +F_{y_2}) - l_r (F_{y_3} + F_{y_4}) \\ \nonumber
	& + & l_w (F_{x_2}+F_{x_4}-F_{x_1}-F_{x_3})
	\end{eqnarray}	

	$F_{x_i}$ and $F_{y_i}$ denote respectively the longitudinal and the lateral tire forces expressed in the vehicle frame; ${F_{aero}=\frac{1}{2}\rho_{air}C_x S V_x^2}$ denote the aerodynamic drag forces with $\rho_{air}$ the mass density of air, $C_x$ the aerodynamic drag coefficient and $S$ the frontal area of the vehicle;  $F_{z_i}$  denote the damped mass/spring forces depending on the suspension travel $\zeta_i$ due to the roll $\theta$ and pitch $\phi$ angles according to Equation~(\ref{eq:veh_dyn_Fz}). The parameters $k_s$ and $d_s$ are respectively the stiffness and the damping coefficients of the suspensions.
	\begin{eqnarray}
		\label{eq:veh_dyn_Fz}
		F_{z_i} = - k_s \zeta_i(\theta, \phi) - d_s \dot{\zeta_i}(\theta, \phi)
	\end{eqnarray}
	
	The position $(X,Y)$ of the vehicle in the ground frame can then be derived using Equations~(\ref{eq:veh_dyn_X}) and (\ref{eq:veh_dyn_Y}).
	\begin{eqnarray}
	\label{eq:veh_dyn_X}
	\dot{X} & = & V_x \cos\psi - V_y \sin\psi\\
	\label{eq:veh_dyn_Y}
	\dot{Y} & = & V_x \sin\psi + V_y \cos\psi	
	\end{eqnarray}
\end{subequations}
	
	\subsection{Wheel dynamics}
	\label{ssec:WheelModel}
	The dynamics of each wheel $i=1..4$ expressed in the pneumatic frame is given by Equation~(\ref{eq:wheel_dyn}):
	\begin{eqnarray} 
		\label{eq:wheel_dyn}
		I_r \dot{\omega}_i & = & T_{{\omega}_i}-r_{eff} F_{xp_i}
	\end{eqnarray}

\subsection{Tire dynamics}
\label{ssec:TireModel}
The longitudinal force $F_{xp_i}$ and the lateral force $F_{yp_i}$ applied by the road on each tire $i$ and expressed in the pneumatic frame are functions of the longitudinal slip ratio $\tau_{x_i}$, the side-slip angle $\alpha_i$, the normal reaction force $F_{z_i}$ and the road friction coefficient $\mu$: 
\begin{subequations}
	\begin{eqnarray}
	F_{xp_i} & =  & f_x(\tau_{x_i}, \alpha_i, F_{z_i}, \mu)\\
	F_{yp_i} & =  & f_y(\alpha_i, \tau_{x_i}, F_{z_i}, \mu)
	\end{eqnarray}
\end{subequations}

The longitudinal slip ratio of the wheel $i$ is defined as following:
\begin{eqnarray}
	\label{eq:slip_ratio}
	\;\tau_{x_i} = \left\{
	\begin{array}{ll}
	\frac{r_{eff} \omega_i - V_{xpi}}{r_{eff}|\omega_i|} & $if $ r_{eff}\omega_i \geq V_{xp_i} $ (Traction phase)$\\
	\frac{r_{eff} \omega_i - V_{xpi}}{|V_{xpi}|} & $if $ r_{eff}\omega_i < V_{xpi} $ (Braking phase)$ 
	\end{array}
	\right.
\end{eqnarray}

The lateral slip-angle $\alpha_i$ of tire $i$ is the angle between the direction given by the orientation of the wheel and the direction of the velocity of the wheel (see Figure~\ref{fig:carSim}):
\begin{eqnarray}
\small
\alpha_f = \delta - \atan \left(\frac{V_y + l_f \dot{\psi}}{V_x \pm l_w \dot{\psi}}\right) ; \; \alpha_r = - \atan \left (\frac{V_y - l_r \dot{\psi}}{V_x \pm l_w \dot{\psi}}\right)
\end{eqnarray}

In order to model the functions $f_x$ and $f_y$, we used the combined slip tire model presented by Pacejka in \cite{Pacejka2002} (cf. Equations (4.E1) to (4.E67)) which takes into account the interaction between the longitudinal and lateral slips on the force generation. Therefore, the friction circle due to the laws of friction (see Equation (\ref{eq:friction_circle})) is respected. Finally, the impact of load transfer between tires is also taken into account through $F_z$. 
\begin{eqnarray}
\label{eq:friction_circle}
||\vec{F}_{xp}+\vec{F}_{yp}|| \leq \mu ||\vec{F}_z||
\end{eqnarray}  

Lastly, the relationships between the tire forces expressed in the vehicle frame $F_x$ and $F_y$ and the ones expressed in the pneumatic frame $F_{xp}$ and $F_{yp}$ are given in Equation~(\ref{eq:frame_F_change}):
\begin{subequations}
	\label{eq:frame_F_change}
	\begin{eqnarray}
	\small{F_{x_i}} & = & \small{(F_{xp_i}\cos\delta_i-F_{yp_i}\sin\delta_i)\cos\phi-F_{z_i}\sin\phi}\\
	\small{F_{y_i}} & = & \small{(F_{xp_i}\cos\delta_i-F_{yp_i}\sin\delta_i)\sin\theta\sin\phi}\\ \nonumber
	& + & \small{(F_{yp_i}\cos\delta_i+F_{xp_i}\sin\delta_i)\cos\theta + F_{z_i}\sin\theta\cos\phi}		
	\end{eqnarray}
\end{subequations}

More details on vehicle dynamics can be found in \cite{Gillespie1997} and \cite{Rajamani2012}.

\section{DEEP LEARNING MODELS}
\label{sec:DL_models}

We propose two different artificial neural network architectures to learn the inverse dynamics of a vehicle, in particular the coupled longitudinal and lateral dynamics. 
An artificial neural network is a network of simple functions called neurons. Each neuron computes an internal state (activation) depending on the input it receives and a set of trainable parameters, and returns an output depending on the input and the activation. Most neural networks are organized into groups of units called layers and arranged in a tree-like structure, where the output of a layer is used as input for the following one. The training of the neural network consists in finding the set of parameters (weights and biases) minimizing the error (or \emph{loss}) between predicted and actual values on a training dataset. In this paper, this training dataset is computed using the 9 DoF vehicle model presented in Section~\ref{sec:vehicle_model}.

\subsection{Dataset}
\label{ssec:dataset}
 
The dataset generated by the 9DoF vehicle model has a total of 43241 instances: it is divided into a train set of 28539 instances and a test set of 14702 instances. The following procedure was used to generate each instance: 

First, a control $u$ to apply is generated randomly, as well as an initial state $\xi^{(0)}$ of the vehicle.
More precisely, the vehicle is chosen to be either in an acceleration phase or in a deceleration phase with equiprobability. 
In the first case, the torques at the front wheels $T_{\omega_1}$ and $T_{\omega_2}$ are set equal to each other and drawn from a uniform distribution between $0$Nm and $750$Nm, while the torques at the rear wheels $T_{\omega_3}$ and $T_{\omega_4}$ are set equal to zero (the vehicle is assumed to be a front-wheel drive one). In the second case, the torques of each wheel are set equal to each other and drawn from a uniform distribution between $-1250$Nm and $0$Nm. In both cases, the steering angle $\delta$ is drawn from a uniform distribution between $-0.5$ and $+0.5$rad.
The initial state $\xi^{(0)}$ is composed of the initial position $(X^{(0)},Y^{(0)})$ of the vehicle in the ground frame, the longitudinal and lateral velocities $V_x^{(0)}$ and $V_y^{(0)}$, the roll, pitch and yaw angles and their derivatives, and the rotational speed $\omega_i^{(0)}$ of the each wheels. The initial longitudinal speed $V_x^{(0)}$ is drawn from a uniform distribution between $5$ and $40$m.s$^{-1}$; the initial lateral speed $V_y^{(0)}$ is drawn from a uniform distribution whose parameters depend of $V_x^{(0)}$; the rotational speed $\omega_i^{(0)}$ is chosen such that the longitudinal slip ratio is zero. All the other initial states are set to zero. 

Secondly, the 9~DoF vehicle model is run for $3$s, starting from the initial state $\xi^{(0)}$ and keeping the control $u$ constant during the whole simulation. 

The resulting trajectories are downsampled to 301 timesteps, corresponding to a sampling time of $10$ms.

Consequently, each instance of the dataset consists in: an initial state $\xi^{(0)}$ of the vehicle, a control $(T_{\omega_1}, T_{\omega_2}, T_{\omega_3}, T_{\omega_4}, \delta)$ kept constant over time, and the associated trajectory obtained $(X^{(0)}, Y^{(0)}), \ldots, (X^{(300)}, Y^{(300)})$.
The dataset generation method is summarized in Algorithm \ref{algo:dataset_gen_algo}.

\begin{algorithm}
\caption{Dataset Generation}
\begin{algorithmic}[1]
\Function{generate instance}{}
   \State $\mathtt{is\_accelerating} \sim {\mathcal {B}}(0, 1)$  \Comment{Coin flipping}
   
   \If{$\mathtt{is\_accelerating} = 1$}
       \State $u_1 \sim {\mathcal {U}}(0, 750)$ \Comment{uniform; in N.m}
       \State $\delta \sim {\mathcal {U}}(-0.5, +0.5)$ \Comment{uniform; in rad}
       \State $u \gets [u_1, u_1, 0, 0, \delta]$
   \ElsIf{$\mathtt{is\_accelerating} = 0$}
       \State $u_1 \sim {\mathcal {U}}(-1250, 0)$ \Comment{uniform; in N.m}
       \State $\delta \sim {\mathcal {U}}(-0.5, +0.5)$ \Comment{uniform; in rad}
       \State $u \gets [u_1, u_1, u_1, u_1, \delta]$
   \EndIf

   \State $V_x^{(0)} \sim {\mathcal {U}}(5, 40)$ \Comment{uniform; in m.s$^{-1}$}
   \State $V_y^{(0)} \sim {\mathcal {U}}(a, b)$ \Comment{uniform; in m.s$^{-1}$}
   \State where $a=\max\left(-1, - \frac{V_x^{(0)}}{3}\right)$
   \State and   $b=\min \left(+1, + \frac{V_x^{(0)}}{3}\right)$

	   \State $\mathtt{trajectory} \gets 9DoF(\xi^{(0)}, u, T_{sim}=3s) \big|_{(X,Y)}$
    \State \textbf{save} $(\xi^{(0)}, u,\ \mathtt{trajectory})$
\EndFunction
\Function{generate dataset}{$n=43241$}
\For{$i\gets 1 \ldots n$}
	\textproc{generate instance()}
\EndFor
\EndFunction
\end{algorithmic}
\label{algo:dataset_gen_algo}
\end{algorithm}

\subsection{Model 1: Multi-Layer Perceptron}
\label{ssec:Model_MLP}

A Multi-Layer Perceptron (MLP), or multi-layer feedforward neural network, is a neural network $f$ whose equations are:
\begin{subequations}
\begin{eqnarray}
\vecbold{h}^{(0)} & = & \vecbold x \\
\vecbold{h}^{(k)} & = &\sigma^{(k)}( \vecbold{W}^{(k)\top} \vecbold{h}^{(k-1)} + \vecbold{b}^{(k)}), \; \text{for } {k=1..L}
\end{eqnarray}
\end{subequations}
where $\vecbold{x}$ denotes the input vector, $\vecbold{h}^{(k)}$ the output of layer $k \in \llbracket 1, L \rrbracket$, $L \in \mathbb N^{*}$ the number of layers of the MLP  and $\sigma^{(k)}$ denotes the $k$-th activation function. $\vecbold{h}^{(L)} = f(\vecbold{x})$ denotes the output vector of the neural network.

\begin{figure}[h!]
	\centering
	\includegraphics[scale=0.6]{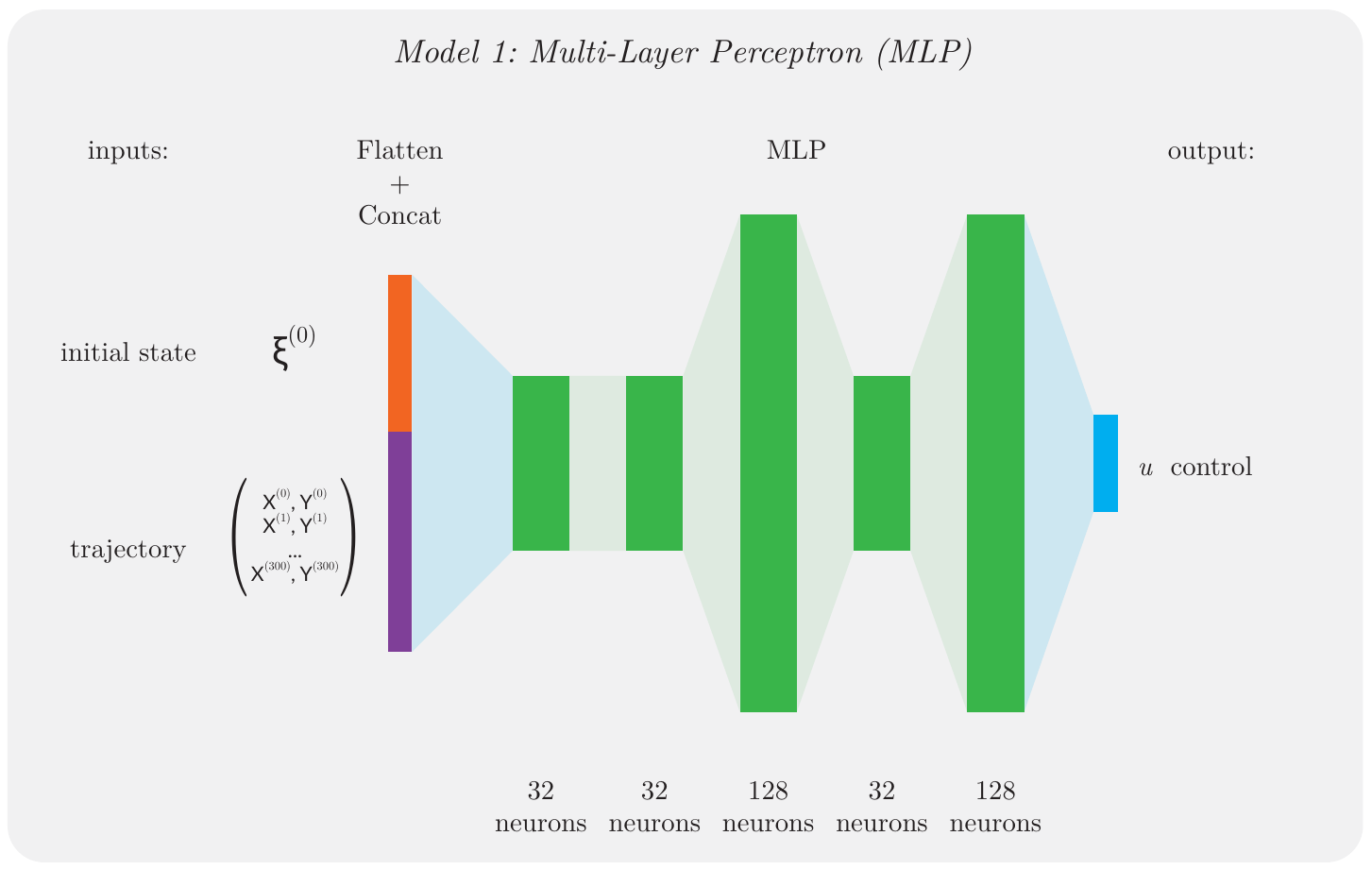}       
	\caption{Multi-Layer Perceptron}
	\label{fig:mlp-full}
\end{figure}

The MLP, presented in Figure~\ref{fig:mlp-full}, is used to predict the constant control $(T_{\omega_1}, T_{\omega_2}, T_{\omega_3}, T_{\omega_4}, \delta)$ to apply given an initial state $\xi^{(0)}$ and a desired trajectory $(X^{(0)}, Y^{(0)}), \ldots, (X^{(300)}, Y^{(300)})$. It is trained on the dataset presented in subection \ref{ssec:dataset}.
It comprises $L=5$ layers, respectively containing 32, 32, 128, 32 and 128 neurons. All the activations functions of the network are rectified linear units (ReLU): ${\sigma(x) = \max(0, x)}$. The loss function used, as well as weights initialization or regularization are discussed in the section~\ref{ssec:Training}, as they are common for the two neural networks proposed. We performed a grid search to choose the sizes of the layers among $3^5=243$ possibilities by allowing each layer to have a size of either 32, 64, or 128 neurons, training the corresponding MLP for 200 epochs and evaluating its performance on the test dataset.

\subsection{Model 2: Convolutional Neural Network}
\label{ssec:Model_CNN}

Convolutional Neural Networks (CNN) are neural networks that use convolution in place of general matrix multiplication in at least one of their layers. A traditional CNN model almost always involves a sequence of convolution and pooling layers.
CNNs have a proven history of being successful for processing data that has a known grid-like topology. For instance, numerous authors make use of CNNs for classification \cite{dong2015vehicle}, or semantic segmentation \cite{badrinarayanan2017segnet} purposes. 

\begin{figure}[h!]
	\centering
	\includegraphics[scale=0.5]{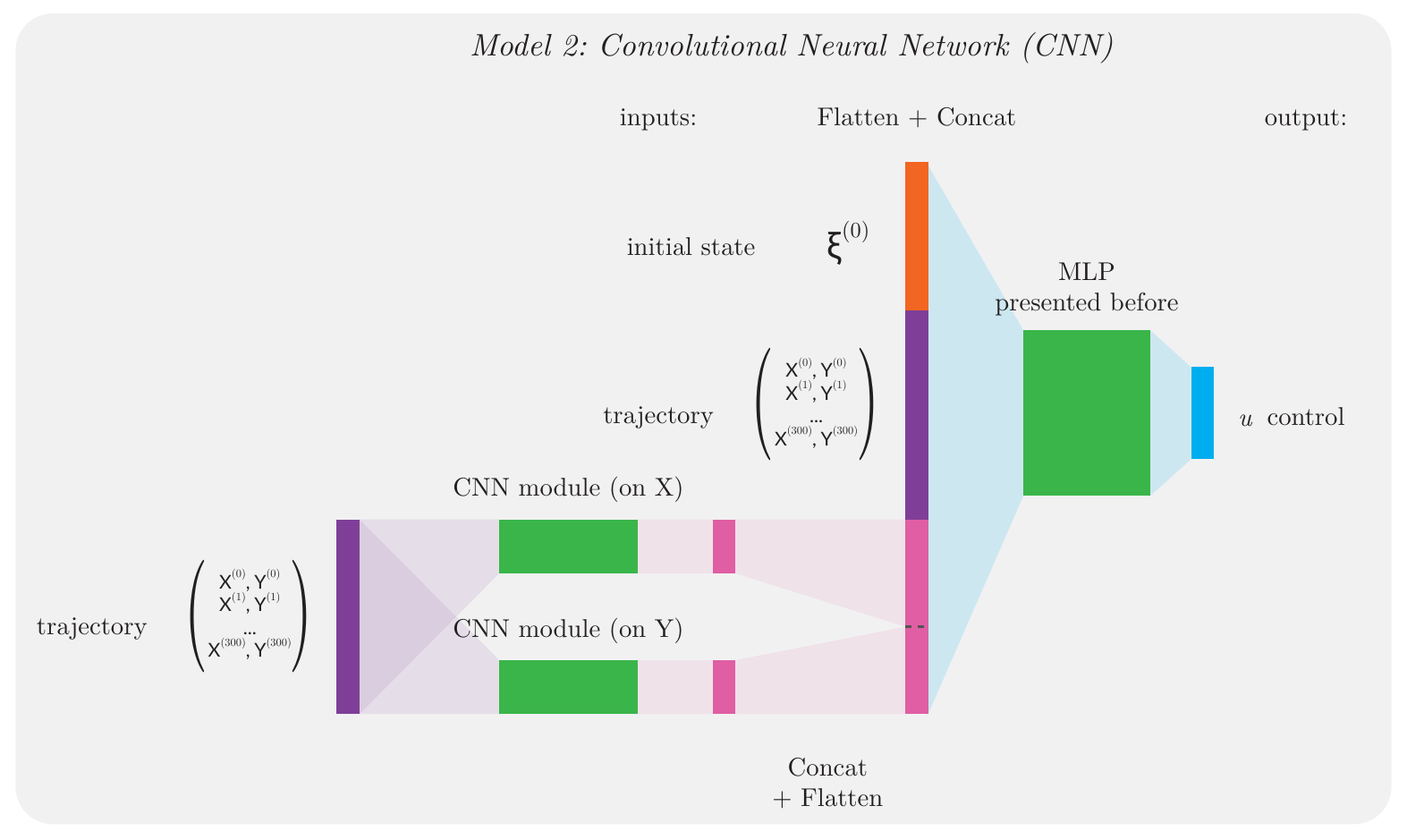}       
	\caption{Convolutional Neural Network}
	\label{fig:cnn-full}
\end{figure}

\begin{figure}[h!]
	\centering
	\includegraphics[scale=0.6]{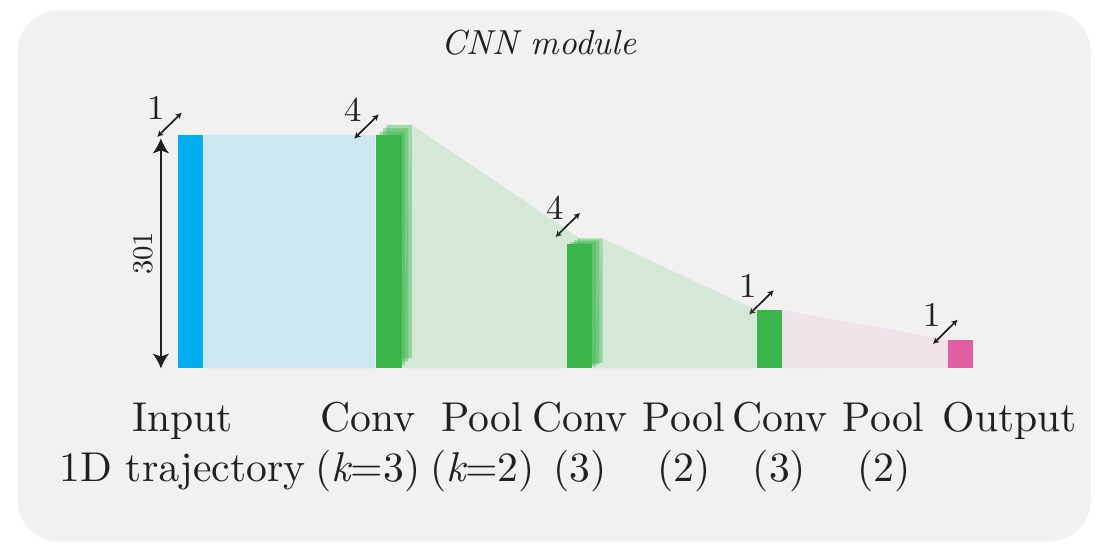}       
	\caption{CNN Module}
	\label{fig:cnn-module}
\end{figure}

We propose to use convolutions to pre-process the vehicle trajectory before feeding it to the MLP, as illustrated in Figure~\ref{fig:cnn-full}. Trajectories are time-series data, which can be thought of as a 1D grid taking samples at regular time intervals, and thus are very good inputs to process with a CNN.
We decided to process the X and Y coordinates separately. 
For each channel $\vecbold{x}$ (either $X$ or $Y$), we construct the following CNN module, which is depicted in Figure~\ref{fig:cnn-module}:
\begin{subequations}
	\begin{eqnarray}
	\vecbold{h}^{(0)} & = & \vecbold{x} \\
	\vecbold{h}^{(k)} & = & \sigma^{(k)}( \pi^{(k)} ( \vecbold{W}^{(k)} * \vecbold{h}^{(k-1)} + \vecbold{b}^{(k)} ) ), \; \text{for } {k=1..L'} \quad \; \;
	\end{eqnarray}
\end{subequations}
where $\vecbold{h}^{(L')}$ is the output of the CNN module, $L' \in \mathbb N^{*}$ the number of layers, $\sigma^{(k)}$ the $k$-th activation function and $\pi^{(k)}$ the $k$-th pooling function. 

The parameters of the CNN module are $L'=3$, with a convolution kernel size of 3 for all convolutions. The activation functions are all ReLU and the pooling functions are all average-pooling of size 2. 
The first two convolutions have 4 feature maps while the last convolution has only 1 feature map. 

As the longitudinal and lateral dynamics are quite different, distinct sets of weights are used for the $X$ and $Y$ convolutions. After processing the X and Y 1D-trajectories by their dedicated CNN module, their output are concatenated.
This new output is then fed to the former MLP whose characteristics remain the same except from the dimension of its input.
The whole model shown in Figure~\ref{fig:cnn-full} is designated as the ``CNN model'' in the rest of this work. 

\subsection{Training procedure}
\label{ssec:Training}

The training procedure is the same for the two neural networks:

	\subsubsection{Weights Initialization \& Batching}
	\label{sssec:weights_initialization}
	
	Each training batch is composed of 32 instances of the dataset.
	The Xavier initialization \cite{Glorot2010} (also known as GLOROT uniform initialization) is used to set the initial random weights for all the weights of our model.

    \subsubsection{Loss function, Regularization \& Optimizer}
    \label{sssec:loss_and_regu_and_optim}
    
    The objective of the training is to reduce the mean square error (MSE) between the controls predicted $u^{pred}$ by the neural network and the ones $u^{real}$ that were really applied to obtain the given trajectory.
    The neural network is trained in order to minimize the loss function $L$ defined by Equation~(\ref{eq:loss_function}) on the train dataset, before evaluation on the test dataset.
	\begin{eqnarray}
		\label{eq:loss_function}
		L = \gamma L_{\delta} + (1 - \gamma) L_{T} + L_{reg}
	\end{eqnarray}
	where 
	\begin{subequations}
	\begin{eqnarray}
		\label{eq:loss_function1}
		L_{\delta}(\delta^{real}, \delta^{pred}) & = & \frac{1}{0.5} MSE(\delta^{real}, \delta^{pred})\\
		\label{eq:loss_function2}
		L_{T}(T_{\omega_i}^{real}, T_{\omega_i}^{pred}) & = & \frac{1}{4 \times 2000} \sum_{i=1}^4 MSE(T_{\omega_i}^{real}, T_{\omega_i}^{pred}) \qquad\\
		\label{eq:loss_function3}
		L_{reg}(W) & = & \gamma_{reg} ||W||_2^2
	\end{eqnarray}
	\end{subequations}
	
	The scaling factors $1/0.5$ and $1/(4\times2000)$ were chosen in order to normalize the steering and the torques. The parameter $\gamma=0.99$ was chosen in order to prioritize the lateral dynamics over the longitudinal one. Equation~(\ref{eq:loss_function3}) is an L2 regularization of our model, where $W$ is the vector containing all the weights of the network. We set $\gamma_{reg} = 10^{-5}$.
	
	To train our model, we used the Adam optimization algorithm \cite{Kingma2014}. It calculates an exponential moving average of the gradient and the squared gradient. For the decay rates of the moving averages, we used the parameters $\beta_1 = 0.9$, $\beta_2 = 0.999$. The values of other parameters were $\alpha = 10^{-3}$ for the learning rate, and $\epsilon = 10^{-8}$ to avoid singular values.

\section{RESULTS}
\label{sec:results}
\begin{figure}[h!]
	\centering
	\includegraphics[]{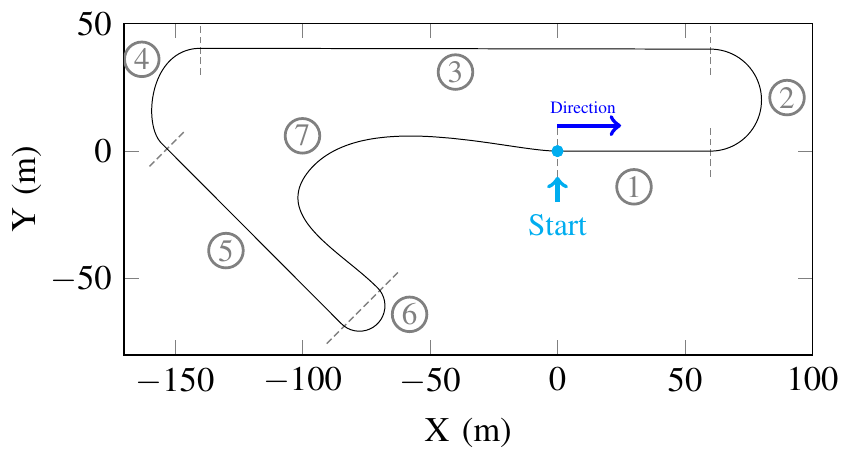}       
	\caption{Top view of the test track; numbers $1$ to $7$ refers to different road sections delimited by dashed lines in order to facilitate the matching with Figures~\ref{fig:steering} to \ref{fig:lateral_error}.}
	\label{fig:track_circuit}
\end{figure}

In order to compare their ability to learn the vehicle dynamics, the two different artificial neural networks are used as ``controllers"\footnote{Properly speaking, they are not real controllers as they to not learn how to reject disturbances and modeling errors.}. The reference track, presented in Figure~\ref{fig:track_circuit}, comprises both long straight lines and narrow curves. The reference speed is set to $V_{ref}=10$m/s on the whole track. 

\subsection{Generating the control commands}
\label{ssec:results_tracking}

\begin{figure}[h!]
	\centering
	\includegraphics[]{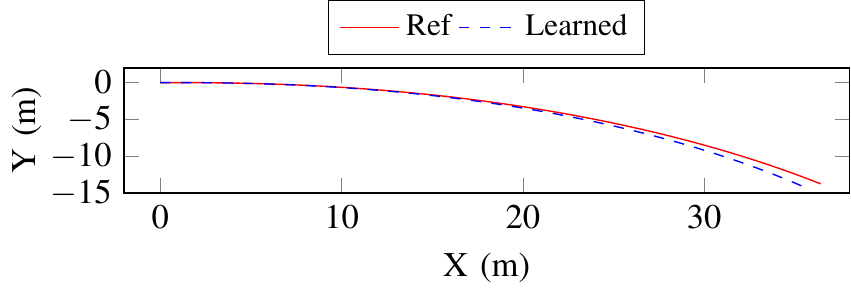}      
	\caption{Example of a training dataset instance: in red, the reference trajectory, in blue the one obtained from the control predicted by the CNN model.}
	\label{fig:ex_trajdataset}
\end{figure}

\begin{figure}[h!]
	\centering
	\includegraphics[]{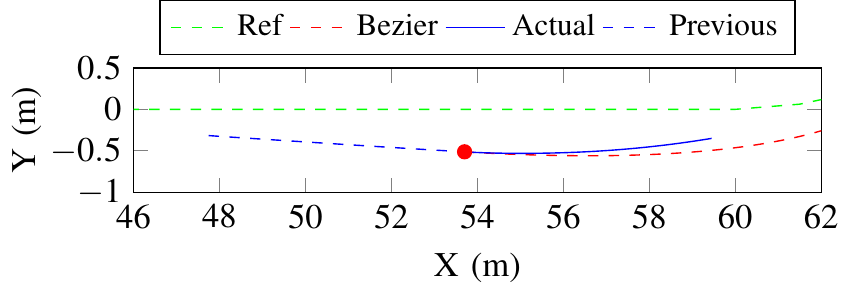}       
	\caption{Example of a Bezier curve (in red) joining the actual position of the vehicle (the red circle) to the reference trajectory (in green). The actual trajectory followed by the vehicle is shown in blue.}
	\label{fig:ex_trajbezier}
\end{figure}

In order to compute the control commands to be applied to the vehicle, the artificial neural network needs to know the trajectory the vehicle has to follow in the next $3$s, as in the train dataset. One problem that arises is that it has only learned to follow trajectories starting from its actual position such as in Figure~\ref{fig:ex_trajdataset}. However, in practice, the vehicle is almost never exactly on the reference path. Therefore, a path section starting from the actual position of the vehicle and bringing it back to the reference path is generated: for that purpose, cubic Bezier curves were chosen as illustrated in Figure~\ref{fig:ex_trajbezier}. 
Thus, at each iteration, 
(i) a Bezier curve with length $3$s is computed to link the actual position of the vehicle to the reference trajectory; 
(ii) a query comprising the previously computed Bezier curve is sent to the artificial neural network;
(iii) the artificial neural network returns the torques at each wheel and the front steering angle to apply
until the next control commands are obtained. The computation sequence is run every $300$ms, even though the query takes less than $2$ms. 

\subsection{Comparison of the models}
\label{ssec:results_comparison}

The results obtained for the MLP and the CNN models are displayed respectively in blue and in red in Figures~\ref{fig:steering} to \ref{fig:lateral_error}. The resulting videos, obtained using the software PreScan \cite{Prescanurl}, are available online\footnote{https://www.youtube.com/watch?v=yyWy1uavlXs}.
Clearly, it appears that the results obtained using a CNN are better than a MLP.
First, we observe that the control commands are smoother in curves with the CNN. There are steep steering (see Figure~\ref{fig:steering}) and front torques (see Figure~\ref{fig:Tfront}) variations for the MLP around $s=360$m  
in road sections n$^\circ4$ and around $s=480$m in road sections n$^\circ6$. In the latter case, the steering angle reaches its saturation value $+0.5$rad and the wheel torques change suddently from $1000$Nm to $-1000$Nm and vice-versa, which is impossible in practice. On the contrary, the control signals of the CNN model remains always smooth and within a reasonable range of values. Secondly, both the longitudinal and lateral errors are smaller for the CNN than the MLP as shown respectively in Table~\ref{tab:perf_longi} and \ref{tab:perf_lateral}.

\begin{table}[h]
	\centering
	\caption{Comparison of the longitudinal performances of the MLP and CNN controllers (in m/s).}
	\label{tab:perf_longi}
	\begin{tabular}{|l|cccc|}
		\hline
		model & RMS  & average & std. dev. & max\\
		\hline
		MLP & 0.76 & -0.29 & 0.70 & -4.94\\
		CNN & 0.60 & -0.39 & 0.46 & -2.33\\
		\hline
	\end{tabular}
\end{table}

\begin{table}[h]
	\centering
	\caption{Comparison of the lateral performances of the MLP and CNN controllers (in m).}
	\label{tab:perf_lateral}
	\begin{tabular}{|l|cccc|}
		\hline
		model & RMS  & average & std. dev. & max\\
		\hline
		MLP & 0.61 & 0.003 & 0.61 & 3.26\\
		CNN & 0.43 & 0.014 & 0.43 & 1.7\\
		\hline
	\end{tabular}
\end{table}

\begin{figure}[h!]
	\centering
	\includegraphics[]{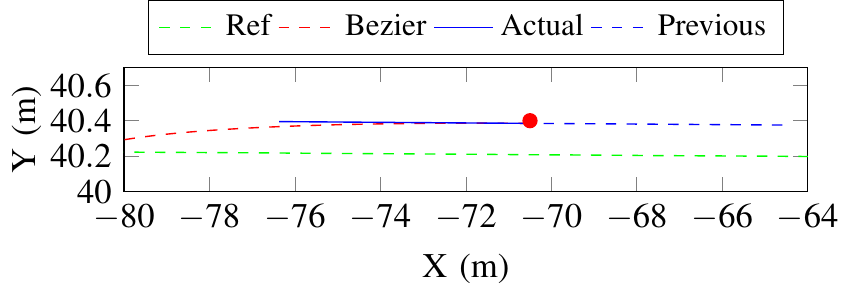}       
	\caption{Example of a Bezier curve on a straight line section of the reference trajectory. The lateral error is not corrected since the convergence of the Bezier curve to the reference trajectory is too slow.}
	\label{fig:CNN_Bezier_static_error}
\end{figure}

However, unlike classic controllers, stability cannot be ensured for these ``controllers" as they are black boxes. In particular, for the CNN, we observe a lateral static error in straight lines. This static error is caused in fact by the Bezier curves which do not converge fast enough to the reference track on straight lines as only the first $300$ms are really followed by the CNN model (see Figure~\ref{fig:CNN_Bezier_static_error}). Moreover, Figure~\ref{fig:steering} shows that the steering angle applied during straight lines is the same for MLP and CNN.

\subsection{Coupling between longitudinal and lateral dynamics}
\label{ssec:results_coupling}

The speed limit a kinematic bicycle model can reach in a curve of radius $R$ is given by Equation~(\ref{eq:V_lim6}) where $\mu=1$ is the road friction coefficient and $g$ the gravity constant \cite{PolackACC2018}. This corresponds to $9.9$m/s ($R=20$m) in road section n$^\circ$2 and $7.0$m/s ($R=10$m) in road section n$^\circ$6. 
As the reference speed is set to $10$m/s throughout the track, conventional decoupled longitudinal and lateral controllers (based on a kinematic bicycle model) will not perform well in road section n$^\circ$6.

\begin{eqnarray}
\label{eq:V_lim6}
V_{kbm_{lim}} & = & \sqrt{0.5 \mu g R}
\end{eqnarray}

On the contrary, both models (especially the CNN) are able to pass this road section, showing the ability of artificial neural networks to handle coupled longitudinal and lateral dynamics. More precisely, we observe in Figure~\ref{fig:V_speed} that the speed is reduced in section n$^\circ6$ because the artificial neural networks deliberately brake (see Figure~\ref{fig:Tfront} and \ref{fig:Trear}), even though the speed of the vehicle is below the reference speed. This is due to the loss function used during training and given by Equation~(\ref{eq:loss_function}) that penalizes more steering angle errors than torque errors. Hence, the models prioritize the lateral over the longitudinal dynamics.
 
Therefore, such ``controllers" are particularly interesting for highly dynamic maneuvers such as emergency situations or aggressive driving where the longitudinal and lateral dynamics are strongly coupled. However, they should be used sparingly as they are only black boxes, or should at least be supervised by model-based systems. Moreover, for normal driving situations, conventional decoupled longitudinal and lateral controller should be preferred.

\newpage

\begin{minipage}[c]{\textwidth}
	\centering
	\includegraphics[]{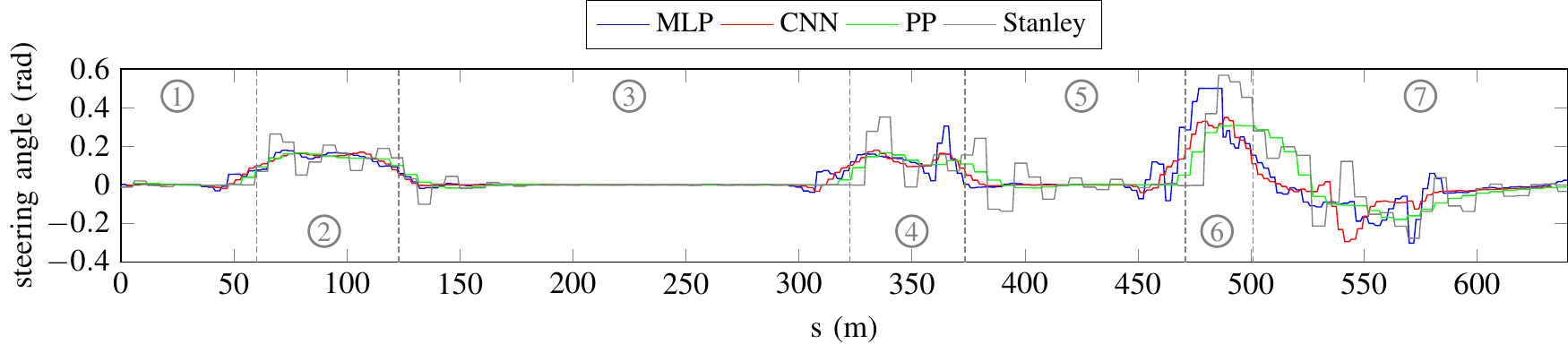}        
	\captionof{figure}{Comparison of the steering command computed by the 
		different controllers. The numbers 1 to 7 correspond to the different road sections presented in Figure~\ref{fig:track_circuit}.}
	\label{fig:steering}
	
	\includegraphics[]{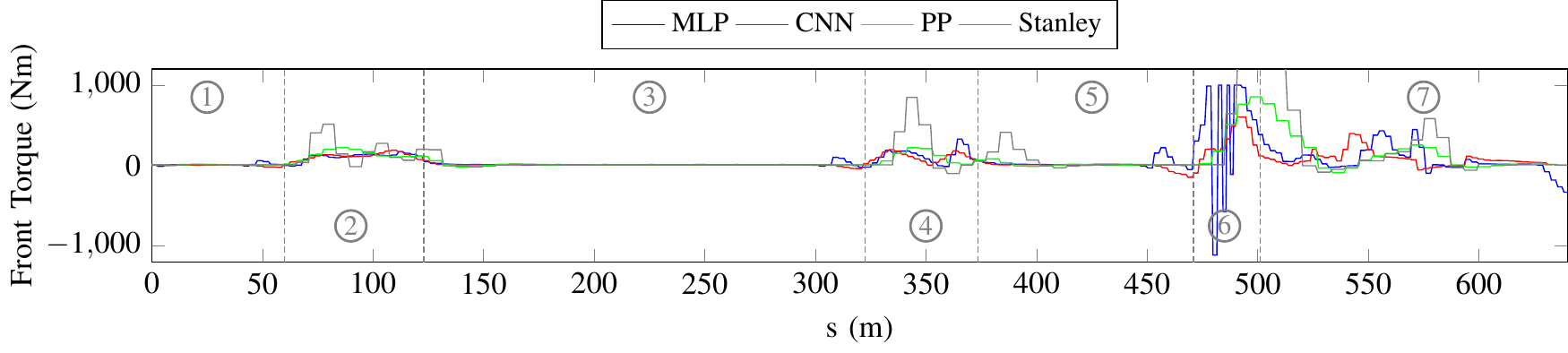}      
	\captionof{figure}{Comparison of the torque applied at the front wheels computed by the different controllers.}
	\label{fig:Tfront}
	
	\includegraphics[]{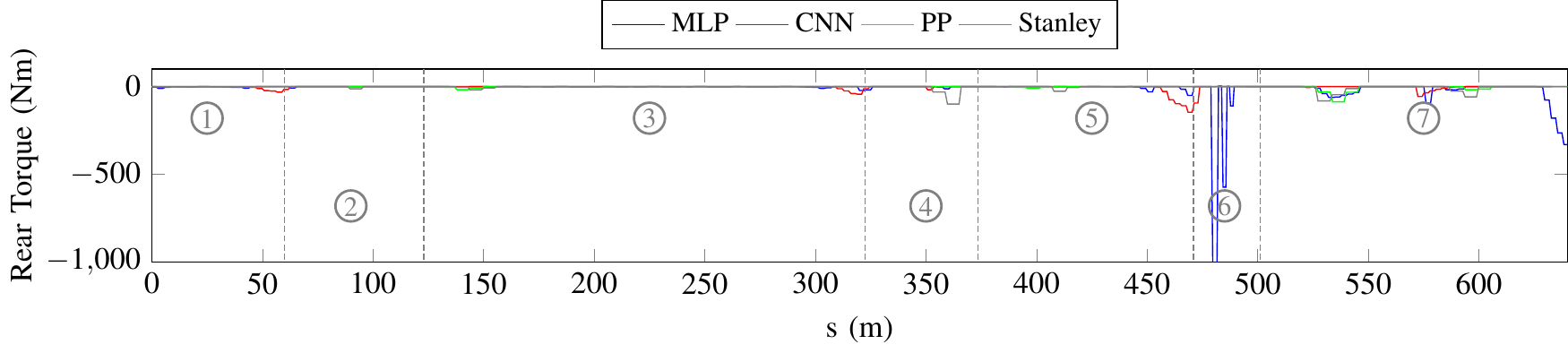} 
	\captionof{figure}{Comparison of the torque applied at the rear wheels computed by the different controllers.}
	\label{fig:Trear}
	
	\includegraphics[]{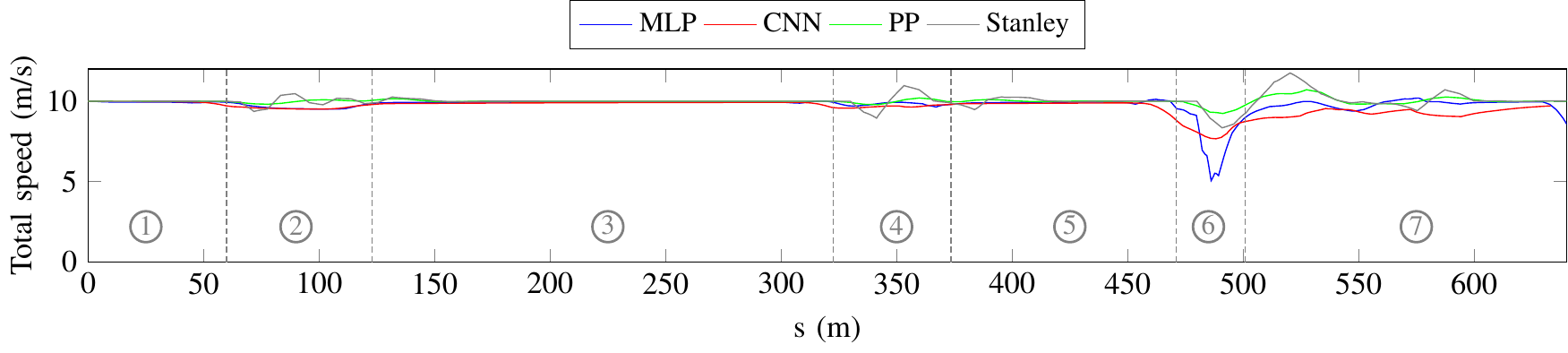}       
	\captionof{figure}{Comparison of the total speed obtained with the different controllers.}
	\label{fig:V_speed}
	
	\includegraphics[]{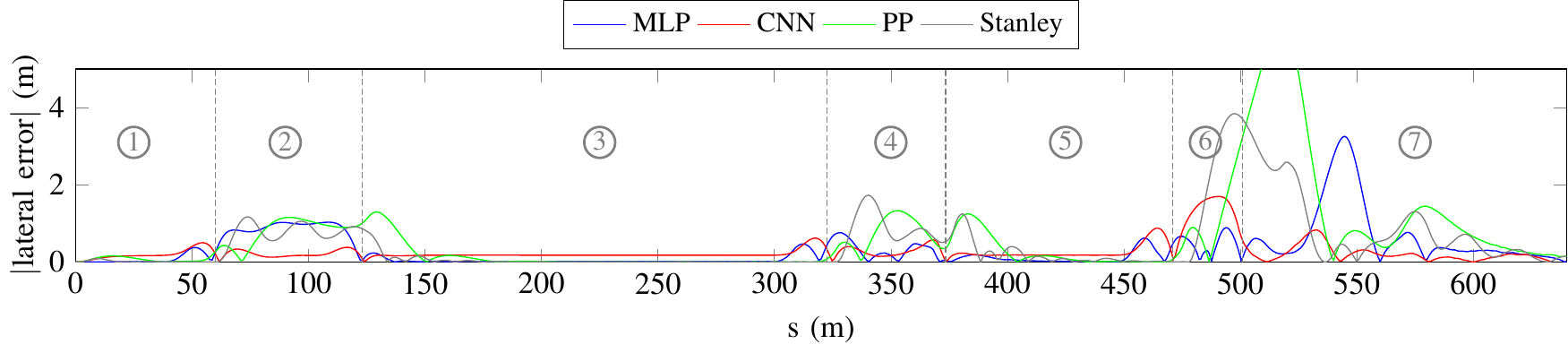}       
	\captionof{figure}{Comparison of the absolute value of the lateral error obtained with the different controllers.
	}
	\label{fig:lateral_error}
\end{minipage}

\clearpage

\subsection{Comparison with decoupled controllers}
\label{ssec:results_decoupled_ctrl_compare}
Finally, the ``controllers" obtained with the MLP and CNN models are compared with commonly used decoupled controllers: the lateral controller is either a pure-pursuit (PP) \cite{Coulter1992} or a Stanley \cite{Thrun2006} controller while in both cases, the longitudinal controller is ensured by a Proportional-Integral (PI) controller with gains $K_P = 600$ and $K_I = 10$. The gain for the front lateral error is $0.75$ for the Stanley controller. The preview distance of the pure-pursuit controller is defined as a function of the total speed $V_g$ at the center of gravity: $L_P = l_f + T_A V_g$ where $T_A = 1.5$s is the anticipation time. The results of the PP and the Stanley controllers are shown respectively in green and grey in Figures~\ref{fig:steering} to \ref{fig:lateral_error}. Clearly, a decrease of performance can be observed when using these decoupled controllers in the challenging part of the track. In particular, the lateral error becomes huge in both cases during the sharp turn of road section n$^\circ6$ while the CNN was able to perform reasonnably well.

\section{CONCLUSIONS}
\label{sec:conclusions}

This work presented some preliminary results on deep learning applied to trajectory tracking for autonomous vehicles. Two different approaches, namely a MLP and a CNN, were trained on a high-fidelity vehicle dynamics model in order to compute simultaneously the torque to apply on each wheel and the front steering angle from a given reference trajectory. It turns out that the CNN model provides better results, both in terms of accuracy and smoothness of the control commands. Moreover, compared to most of the existing controllers, it is able to handle situations with strongly coupled longitudinal and lateral dynamics in a very short time. However, the controller obtained is a black-box and should not be used in standalone.

The results proved the ability of deep learning algorithms to learn the vehicle dynamics characteristics. This opens a wide range of new possible applications of such techniques, for example for generating dynamically feasible trajectories. Future work will focus on (i) replacing the complex dynamics models by a learned off-line model in Model Predictive Control for motion planning, (ii) using Generative Adversarial Networks (GAN) to generate safe trajectories where the learned dynamics is used as constraint, and (iii) performing real-world experiments with our approach on a real car. 

\bibliographystyle{IEEEtran}
\bibliography{arxivdevineauitsc2018}

\addtolength{\textheight}{-12cm}

\end{document}